\documentclass[sigconf]{acmart}
\usepackage{multirow}
\usepackage{balance}
\AtBeginDocument{%
  \providecommand\BibTeX{{%
    \normalfont B\kern-0.5em{\scshape i\kern-0.25em b}\kern-0.8em\TeX}}}

\copyrightyear{2021}
\acmYear{2021}
\setcopyright{acmcopyright}\acmConference[MM '21]{Proceedings of the 29th ACM
International Conference on Multimedia}{October 20--24, 2021}{Virtual Event, China}
\acmBooktitle{Proceedings of the 29th ACM International Conference on Multimedia
(MM '21), October 20--24, 2021, Virtual Event, China}
\acmPrice{15.00}
\acmDOI{10.1145/3474085.3475449}
\acmISBN{978-1-4503-8651-7/21/10}

\begin{document}
\fancyhead{}
\title{Implicit Feature Refinement for Instance Segmentation}

\author{Lufan Ma$^{1\ast\dagger}$,\quad Tiancai Wang$^{2\ast}$,\quad Bin Dong$^{2}$,\quad Jiangpeng Yan$^{1}$,\quad Xiu Li$^{1\ddagger}$,\quad Xiangyu Zhang$^{2}$}

\makeatletter
\def\authornotetext#1{
\if@ACM@anonymous\else
    \g@addto@macro\@authornotes{
    \stepcounter{footnote}\footnotetext{#1}}
\fi}
\makeatother
\authornotetext{$^{\ast}$Equal contribution.$^{\dagger}$Work done during an internship at Megvii Technology. $^{\ddagger}$Corresponding author. This work is supported by the National Key R $\&$ D Plan of the Ministry of Science and Technology (Project No. 2020AAA0104400).}

\affiliation{
 \institution{\textsuperscript{\rm 1}Tsinghua University\city{Beijing}\country{China}}
 \institution{\textsuperscript{\rm 2}MEGVII Technology\city{Beijing}\country{China}}
 }
\email{{malf19,yanjp17}@mails.tsinghua.edu.cn,li.xiu@sz.tsinghua.edu.cn,{wangtiancai,dongbin,zhangxiangyu}@megvii.com}

\def\authors{Lufan Ma, Tiancai Wang, Bin Dong, Jiangpeng Yan, Xiu Li, Xiangyu Zhang}

\renewcommand{\shortauthors}{Lufan Ma, et al.}

\begin{abstract}
  We propose a novel implicit feature refinement module for high-quality instance segmentation. Existing image/video instance segmentation methods rely on explicitly stacked convolutions to refine instance features before the final prediction. In this paper, we first give an empirical comparison of different refinement strategies, which reveals that the widely-used four consecutive convolutions are not necessary. As an alternative, weight-sharing convolution blocks provides competitive performance. When such block is iterated for infinite times, the block output will eventually converge to an equilibrium state. Based on this observation, the implicit feature refinement (IFR) is developed by constructing an implicit function. The equilibrium state of instance features can be obtained by fixed-point iteration via a simulated infinite-depth network. Our IFR enjoys several advantages: 1) simulates an infinite-depth refinement network while only requiring parameters of single residual block; 2) produces high-level equilibrium instance features of global receptive field; 3) serves as a plug-and-play general module easily extended to most object recognition frameworks. Experiments on the COCO and YouTube-VIS benchmarks show that our IFR achieves improved performance on state-of-the-art image/video instance segmentation frameworks, while reducing the parameter burden (\textit{e.g.} 1\% AP improvement on Mask R-CNN with only 30.0\% parameters in mask head). Code is made available at \href{https://github.com/lufanma/IFR.git}{\textit{https://github.com/lufanma/IFR.git}}.
\end{abstract}



\maketitle

\section{Introduction}
\label{introduction}
Owing to the great success of deep convolutional networks~\cite{he2016deep, chen2017deeplab, long2015fully}, object recognition tasks, such as object detection and instance segmentation, have achieved impressive progress during the last decade. Mainstream methods follow the "backbone with head" paradigm~\cite{he2017mask, chen2019hybrid, lin2017focal, tian2019fcos}. The backbone network~\cite{he2016deep}, pretrained on ImageNet, extracts features from input images. Then, the extracted features are fed into the head network to generate predictions for different recognition tasks, \textit{i.e.}, classification, box regression and mask segmentation.

\begin{figure}[t]
  \centering
  \includegraphics[scale=0.50]{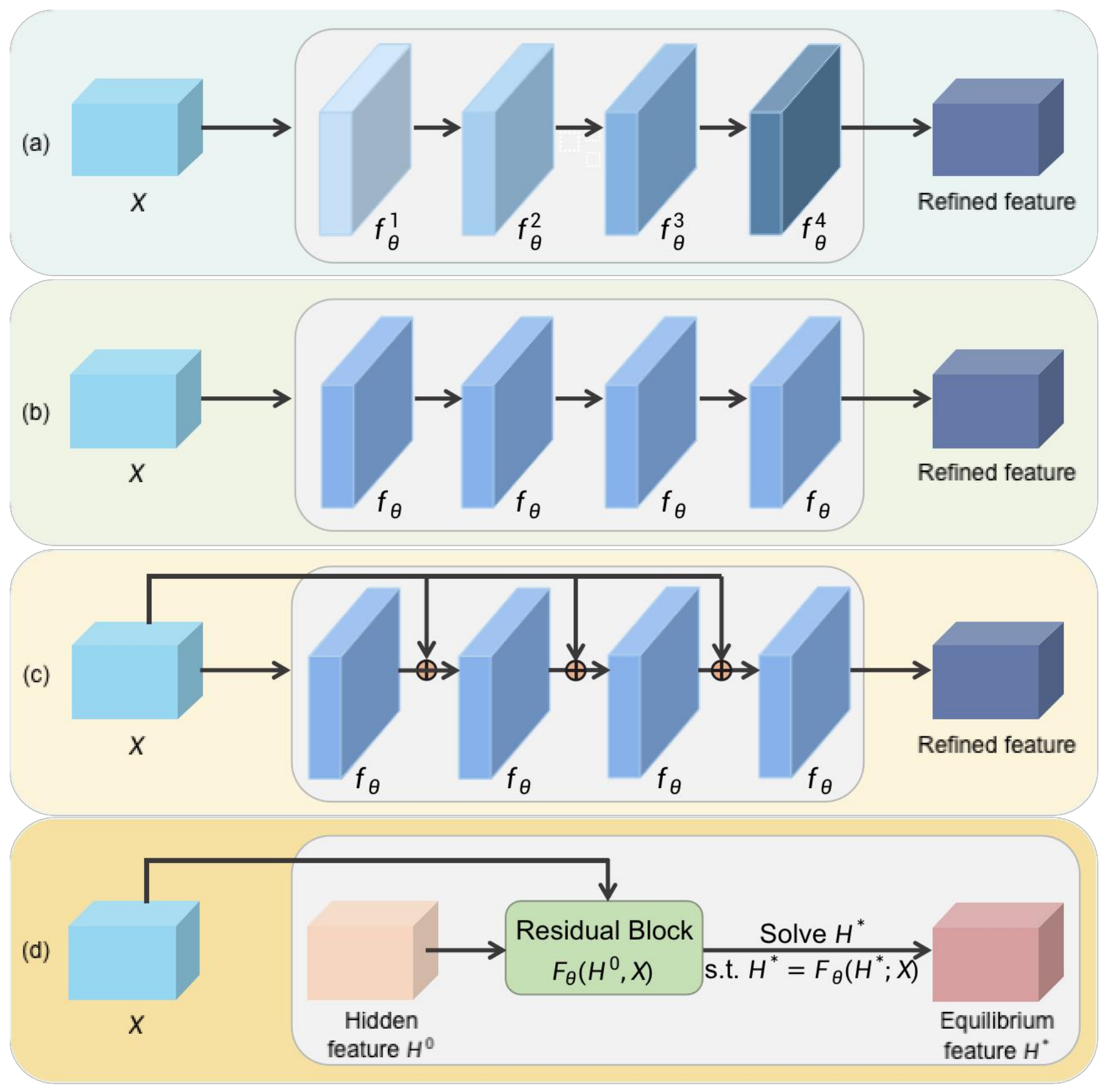}
\caption{Comparison of different feature refinement strategies in the mask head of Mask R-CNN~\cite{he2017mask}. (a) stacks four weight-independent $3\times3$ convolutions ($f_{\theta}^1$-$f_{\theta}^4$). (b) stacks four weight-sharing $3\times3$ convolutions $f_{\theta}$. (c) stacks four weight-sharing convolution block $f_{\theta}$ with skip connection. (d) develops the implicit feature refinement via fixed-point iteration, where a residual block with double residual connections is adopted as the transformation $F_{\theta}$.}
\label{fig1}
\end{figure}

Most image/video instance segmentation~\cite{he2017mask, huang2019mask, yang2019video, cao2020sipmask} and object detection approaches~\cite{lin2017focal, tian2019fcos} share similar design of the head network. Specifically, four consecutive $3\times3$ convolutions are usually stacked in the head network to refine coarse instance features (see Fig.~\ref{fig1}(a)) before the final prediction. Indeed, such explicit refinement design enlarges the receptive field and improves the semantic level. However, simply stacking weight-independent convolutions have some limits. The receptive field obtained is still limited under the explicit setting since it has been proven that the effective receptive field is usually much more smaller compared to theoretical receptive field. Also, such method tends to forget the original instance signal, suffering from the degraded performance and large parameter burden as the number of convolutions increases.

One question may arise: is it possible to replace such stacked $3\times3$ convolutions with a light-weight module while achieving promising performance? One direct solution is to make these four convolutions weight-sharing (see Fig.~\ref{fig1}(b)). To demonstrate it, we first give an empirical comparison with the weight-sharing strategy. We find that the widely-used four weight-independent $3\times3$ convolutions are not necessary in the head network. In contrast, the weight-sharing design can achieve competitive performance with less parameters.

To prevent the weight-sharing network from degraded performance, shortcut connection in ResNet~\cite{he2016deep} can be introduced (see Fig.~\ref{fig1}(c)). The shortcut path can propagate the gradients smoothly and avoid the network forgetting original instance features. While improving the overall segmentation performance, the weight-sharing strategy is still restricted by limited receptive field. Therefore, we explore a new perspective: \textit{What if refining the instance features via a recursive network stacked with infinite blocks?} An intuitive idea is that we stack the weight-sharing blocks for infinite times. In this way, we observe that the output of the refinement network will eventually converge to an equilibrium state as the number of blocks increases, which can be represented as:
\begin{eqnarray}\label{eq1}
H^{\ast}=F_{\theta}(H^{\ast};X),
\end{eqnarray}
where $F_{\theta}$ represents the nonlinear transformation with parameters $\theta$. $H^{\ast}$ and $X$ are the equilibrium instance feature and original coarse instance feature, respectively.

Inspired by DEQ~\cite{bai2019deep, bai2020multiscale}, we introduce the implicit feature refinement (IFR) to enhance feature representation via a simulated infinite-depth network, where the transformation $F_{\theta}$ is employed to build the implicit model. Based on the Eq.~\ref{eq1}, we can tackle the instance feature refinement as a fixed-point iteration problem. For the explicit design of transformation $F_{\theta}$, we simply adopt a single residual block with double residual connections to reduce the overall parameters of mask head (see Fig.~\ref{fig1}(d)). For the forward computation, several black-box Root-Find solvers (\textit{e.g.}, Broyden method~\cite{bai2020multiscale}) can be adopted to solve the root $H^{\ast}$ of the equilibrium equation presented in Eq.~\ref{eq1}. While for the back-propagating, gradients are computed based on the equilibrium feature solved.

Different from the explicitly stacked convolutions, IFR enjoys several advantages: 1) Simulates an infinite-depth feature refinement network only with parameters of a single residual block; 2) Produces the equilibrium instance features of large receptive field, which is beneficial to instance segmentation. It should be noted that IFR can be easily extended to the head network of most image/video instance segmentation frameworks, including both two-stage and one-stage approaches. 

Extensive experiments on the COCO benchmark~\cite{lin2014microsoft} demonstrate that our proposed IFR benefits most image instance segmentation frameworks. For example, it improves the segmentation AP by 1.0\% on Mask R-CNN~\cite{he2017mask} with ResNet-50-FPN backbone. Furthermore, it is also robust and flexible for video instance segmentation on YouTube-VIS~\cite{yang2019video} benchmark, achieving promising improvements on the state-of-the-art frameworks. Also, it can be applied to several one-stage object detectors with slight modifications.

Our main contributions are summarized as:
\begin{itemize}
  \item Empirical experiments are conducted to analyze the effect of multiple stacked convolutions on refining the instance features.
  \item Implicit feature refinement (IFR), which only requires parameters of a residual block, is proposed to produce high-level equilibrium instance features of global receptive field by fixed-point iteration.
  \item Double residual network is further introduced to prevent the network from degraded performance and propagate the gradients smoothly.
  \item The proposed IFR is exactly a general plug-and-play module that can be easily extended to most instance segmentation frameworks and serves as a strong alternative of explicitly stacked convolutions in the head network.
\end{itemize}

\section{Related Work}
\subsection{Implicit Modeling}
The implicit models aim to solve a fixed-point equation of hidden states in a recursive feedforward neural networks~\cite{el2019implicit}. Recurrent back-propagation algorithms~\cite{almeida1990learning, liao2018reviving, pineda1987generalization} utilized implicit differentiation techniques to train recurrent systems. Recently, implicit models have attracted attention in network design. Neural ODEs~\cite{chen2018neural, haber2017stable} implicitly models a recursive residual block with ODE solvers to simulate an infinite-depth ResNet, while~\cite{simard1988fixed} provides a systematic analysis of the stabilities properties in the recurrent backpropagation algorithm. For sequence modeling, DEQ~\cite{bai2019deep} employs black-box root solvers to find the fixed point equilibrium, while TrellisNet~\cite{bai2018trellis} designs truncated recurrent networks in a weight-tied way. Analogously, RAFT~\cite{teed2020raft} iteratively updates the fixed flow field through a lot of modified GRU units. Based on~\cite{bai2019deep}, Bai further develops a multiscale deep equilibrium model (MDEQ)~\cite{bai2020multiscale} for both image classification and semantic segmentation while i-FPN~\cite{wang2020implicit} presents an implicit feature pyramid network for object detection. 

\subsection{Instance Segmentation}
\textbf{Image Instance Segmentation:} Existing methods can be divided into two groups. Two-stage methods~\cite{he2017mask, huang2019mask, chen2019hybrid, kirillov2020pointrend, cheng2020boundary, kim2021devil, ma2021matting} follow the “detect-then-segment” paradigm. Typically, Mask R-CNN~\cite{he2017mask} extends Faster R-CNN~\cite{ren2015faster} by adding a FCN mask branch. Based on~\cite{he2017mask}, Mask Scoring R-CNN~\cite{huang2019mask} further addresses the misalignment problem between the mask quality and classification score. HTC~\cite{chen2019hybrid} interweaves the box and mask branches in a multi-stage cascaded manner. PointRend~\cite{kirillov2020pointrend} refines the boundary details by adaptively sampling points. On the other hand, one-stage methods~\cite{xie2020polarmask, wang2020solo, wang2020solov2, tian2020conditional, luo2021coarse} incorporate the mask prediction into simple FCN-like framework without RoI cropping. PolarMask~\cite{xie2020polarmask} encodes the instance mask by the instance contour in polar coordinates. SOLO~\cite{wang2020solo} directly outputs full instance masks by locations without detection while CondInst~\cite{tian2020conditional} introduces the conditional convolutions to predict the instance-aware masks.

\noindent \textbf{Video instance segmentation:} VIS requires simultaneous detection, segmentation, and tracking of instances across frames in videos. MaskTrack R-CNN~\cite{yang2019video} extends the Mask R-CNN~\cite{he2017mask} with a new tracking branch and external memory that saves the features of instances across multiple frames. Based on HTC~\cite{chen2019hybrid}, Maskprop~\cite{bertasius2020classifying} re-uses the predicted masks to crop the extracted features, then propagates them temporally to improve the segmentation and tracking. STEm-Seg~\cite{Athar_Mahadevan20ECCV} proposes to model video clips as 3D space-time volumes and then separates object instances by clustering learned embeddings. Wang et al.~\cite{wang2020end} proposes an end-to-end framework VisTR built upon Transformers. In this paper, we present the implicit feature refinement to produce high-level segmentation representations for both two-stage and one-stage methods.

\subsection{Object Detection}
Existing object detection approaches can also be summarized into two categories. Two-stage detectors~\cite{ren2015faster, cai2018cascade, singh2018analysis} first generate the foreground proposals by region proposal network (RPN). The proposals generated are further classified and regressed in the second stage. In contrast, one-stage detectors~\cite{liu2016ssd, lin2017focal, tian2019fcos, duan2019centernet, zhang2020bridging, bochkovskiy2020yolov4, redmon2016you} directly predicts the locations and categories of objects in a single-shot manner.  Among these methods, anchor-based frameworks, such as SSD~\cite{liu2016ssd}, YOLO~\cite{redmon2016you, bochkovskiy2020yolov4}, RetinaNet~\cite{lin2017focal} perform the predictions on pre-defined anchors, which densely cover the spatial positions. Anchor-free approaches~\cite{tian2019fcos, duan2019centernet, zhang2020bridging} replace the hand-crafted anchors by reference points. Recently, end-to-end detectors~\cite{carion2020end, zhu2020deformable, sun2020sparse} remove the hand-crafted anchors and non-maximum suppression via bipartite matching. The implicit feature refinement introduced in this paper can be used to refine the instance features of one-stage object detectors as well.

\section{Methodology}
In this section, we first provide analysis on the explicit feature refinement with multiple stacked convolutions. Then the implicit feature refinement that implicitly refines the instance features for instance segmentation is described. Double residual network (Sec.~\ref{double_residual_net}) and hybrid optimization (Sec.~\ref{hybrid_optimization}) employed in the implicit module are further introduced. Finally, we further extend the implicit refinement module to the head network of one-stage object detectors.

\subsection{Analysis on Explicit Feature Refinement}
For two-stage instance segmentation, existing methods~\cite{he2017mask, huang2019mask, kirillov2020pointrend} leverage multiple $3\times 3$ convolutions to refine RoI features in the head network (\textit{e.g.}, four $3\times 3$ convolutions with 256 channels in Mask R-CNN). Formally, such explicit feature refinement can be represented as:
\begin{eqnarray}\label{eq2}
H=f_{\theta}^M(\cdots f_{\theta}^2(f_{\theta}^1(X))),
\end{eqnarray}
where $X\in R^{14\times 14\times 256}$ is the RoI feature cropped by RoIAlign. $M$ is the total number of stacked convolutions. $f_{\theta}^i$ represents the $i_{th}$ $3\times 3$ convolution with parameter $\theta$. $H$ is the output feature refined by stacked convolutions.

To demonstrate the effect of stacked convolutions on feature refinement, we perform a simple ablation study on the number of stacked convolutions on the COCO~\cite{lin2014microsoft} benchmark. As shown in Tab.~\ref{Tab1}, the experiments are conducted on Mask R-CNN~\cite{he2017mask} with ResNet-50 backbone and tested on COCO \textit{val2017} set. The ablation results show continuous improvements as the number of stacked convolutions increases.  Therefore, such stacked convolutions improve the segmentation performance due to the enlarged receptive field and semantic levels of RoI features, revealing their superiority on feature refinement. For more details, please kindly refer to the visualization in Fig.~\ref{fig4}.

However, these stacked convolutions lead to large parameter burden simultaneously. As mentioned in Sec.~\ref{introduction}, one direct way to reduce the overall parameters is to share the weights across the stacked convolutions. Tab.~\ref{Tab1} also reveals that weight-sharing convolutions can achieve the same segmentation performance compared to the weight-independent counterpart. Therefore, the stacked four $3\times 3$ convolutions are not necessary for the mask head network. Instead, a single weight-sharing block enables the same performance with the parameters of one convolution layer.

\begin{table}[h]
 \caption{The effect of stacked $3\times3$ convolutions in the mask head of Mask R-CNN~\cite{he2017mask} on segmentation performance. $^\ast$ indicates the weight-sharing version of stacked convolutions and achieves the same performance compared to the weight-independent counterpart.}\label{Tab1}
  \begin{tabular}{c|ccccc|c}
    \toprule
    Num. of convs & 0 & 1 & 2 & 3 & 4 & 4$^\ast$  \\
    \midrule
    mAP & 30.4 & 32.7 & 34.0 & 34.8 &\textbf{35.0} & \textbf{35.0}\\
    \bottomrule
  \end{tabular}
\end{table}

\begin{figure*}[hbt]
\centering
\includegraphics[scale=0.52]{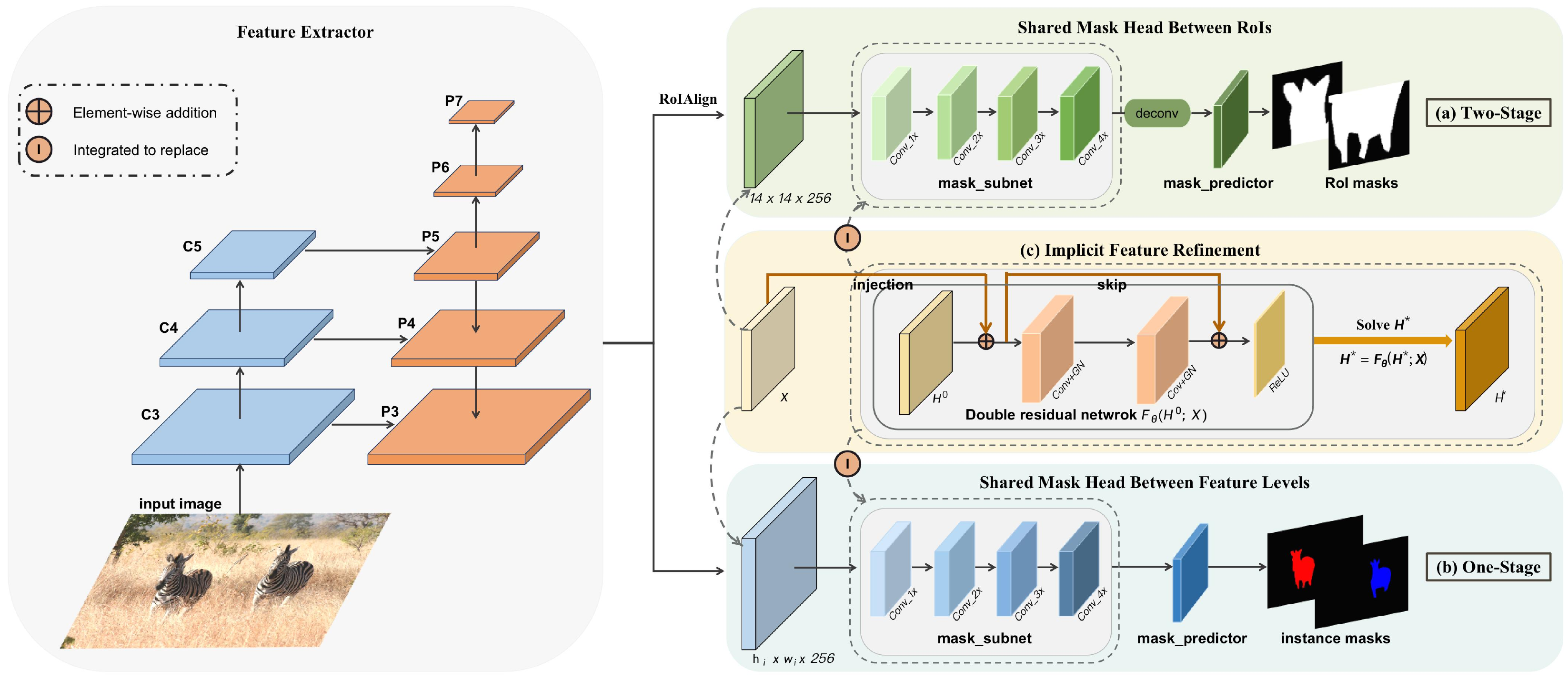}
\caption{The overall architectures of one stage and two-stage instance segmentation with the proposed implicit feature refinement. Note that the implicit refinement module contains only one residual block with double residual connections.}     \label{fig2}
\end{figure*}

\subsection{Implicit Feature Refinement}
As depicted in Fig.~\ref{fig2}, the image $I$ is first fed to the feature extractor, which includes backbone (\textit{e.g.} ResNet-50~\cite{he2016deep} ) and FPN~\cite{lin2017feature}, to produce the pyramidal multi-scale features $\{P3, P4, ..., P_{7}\}$.
For two-stage methods (\textit{e.g.} Mask R-CNN~\cite{he2017mask}), bounding boxes generated from the detection branch (ignored in the Fig.~\ref{fig2}) together with the corresponding pyramid feature $P_{i}$ perform the RoIAlign to generate the fixed-size (\textit{e.g.} $14\times14$) instance features for the mask branch (see Fig.~\ref{fig2}(a)). For one-stage approaches (\textit{e.g.} CondInst~\cite{tian2020conditional}), the mask branch tends to employ the whole feature of high resolution level $P_{3}$ to directly produce the full masks without RoIAlign (see Fig.~\ref{fig2}(b)). For the mask head network in both one-stage and two-stage frameworks, the RoI feature or the whole feature will pass through multiple stacked $3\times3$ convolutions for feature refinement. Afterwards, the refined features are used to produce the mask predictions by specific mask predictors. The proposed implicit feature refinement module can be integrated into both two-stage and one-stage instance segmentation frameworks to replace the stacked convolutions in the mask head.

The implicit feature refinement module (see Fig.~\ref{fig2}(c)) has two inputs: instance feature (or whole feature) $X$ and the zero-initialized hidden feature $H^{0}$. 
Taking the two-stage Mask R-CNN as an example, the unrolling process of implicit refinement module can be formulated as:
\begin{eqnarray}\label{eq3}
H^{i+1}=F_{\theta}(H^{i};X), {\,i\,=\,0,\,1,\cdots,\,N-1},
\end{eqnarray}
where $X\in R^{14\times14\times256}$ represents the instance feature cropped by RoIAlign. $F_{\theta}$ is the transformation block, which is shared for all stacked blocks. $H^{i+1}$ denotes the output of the $i_{th}$ transformation block. $N$ is the overall number of stacked transformation blocks.

When such unrolling process is repeated for infinite times, the IFR module exactly simulates an infinite-depth network. Intuitively, as we perform the transformation $F_{\theta}$ for infinite times ($N\to+\infty$), the network output will eventually reach the equilibrium state as formulated in Eq.~\ref{eq4}.
\begin{eqnarray}\label{eq4}
{\lim_{i\to+\infty}F_{\theta}(H^{i};X)=F_{\theta}(H^{*};X)=H^{*}},
\end{eqnarray}
where $H^{\ast}$ is the equilibrium state of hidden instance feature. Based on this analysis, the optimal instance feature predicted by our implicit module is exactly equivalent to the fixed point $H^{\ast}$ of Eq.~\ref{eq4}. Besides, the fixed point can be directly solved by several black-box Root-Find solvers rather than explicitly unrolling the transformation blocks.

\subsection{Double Residual Network}
\label{double_residual_net}
Intuitively, the transformation block $F_{\theta}$ is a key component since it is used to construct the implicit function that interacts the instance feature with hidden feature. Therefore, the explicit design of $F_{\theta}$ determines the capability of feature refinement. 

Inspired by~\cite{he2016deep}, we introduce a simple yet effective strategy, called double residual connection, for the transformation $F_{\theta}$. Double residual connections are designed to avoid gradient vanishing of the simulated infinite-depth network. Its explicit design is illustrated in Fig.~\ref{fig2}. The input feature $X$ and the initial hidden feature $H^{i}$ are first added. Then the summed features of $X$ and $H^{0}$ pass through through a well-designed residual block~\cite{he2016deep} with residual connection. Note that the addition between $X$ and $H^{0}$ perform another residual connection implicitly. Overall, the double residual network $F_{\theta}$ can be formally expressed as:
\begin{eqnarray}\label{eq5}
F_{\theta}(H^{i};X)=W_{2}(\sigma(W_{1}(R))) +W_{s}(R),
\end{eqnarray}
where $\sigma$ is the activation function ReLU. $W_{s}$ performs a linear identity mapping in the shortcut path. $W1$ and $W2$ denote an integrated layer consisting of a $3\times3$ convolution and normalization, where group normalization (GN)~\cite{wu2018group} is adopted for stable training. $R=H^{0}+X$ is the summed features of $X$ and $H^{0}$.

To this end, we can reformulate the feature refinement as implicit residual learning with respect to both the instance feature $X$ and the summed feature $R$. Owing to the advantage of residual learning, our double residual connection strategy effectively eases the training of network, thus enhancing the feature refinement for instance segmentation. Note that our IFR module only contains parameters of a single residual block, which are much fewer compared to the stacked weight-independent convolutions.

\subsection{Hybrid Optimization}
\label{hybrid_optimization}
In this section, we describe the hybrid optimization process of the instance segmentation framework with the proposed IFR.

\noindent \textbf{Explicit Optimization:}
For the instance segmentation framework with implicit feature refinement, only the IFR part obtains the optimal features based on the fixed-point iteration. Therefore, the rest networks (\textit{i.e.}, feature extractor, mask predictor) still follow the explicit optimization. The explicit optimization includes the forward process by the explicit network and backward propagation via the chain rule.

\noindent \textbf{Implicit Optimization:}
The equilibrium state presented in Eq.~\ref{eq4} can be directly reformulated as a root-finding problem:
\begin{eqnarray}\label{eq6}
{\Phi}_{\theta}(H^{\ast};X)=F_{\theta}(H^{\ast};X)-H^{\ast}=0,
\end{eqnarray}
where $H^{\ast}$ is the root that satisfies $\Phi_{\theta}(H^{\ast};X)=0$.
Moreover, it is essentially the fixed point that represents the equilibrium hidden feature to be solved.

For the forward computation, several black-box root-solvers (\textit{e.g.}, Newton, quasi-Newton methods) can be employed to find the root $H^{\ast}$~\cite{bai2020multiscale}. While for the backward propagation, we follow~\cite{bai2019deep} and leverage the equilibrium hidden feature $H^{\ast}$ solved above to compute the gradients of both the parameters $\theta$ of $F_{\theta}$ and the input feature $X$. Consider the segmentation loss as:
\begin{eqnarray}\label{eq7}
L(\hat{m},m)=L(f(H^{\ast}),m),
\end{eqnarray}
where $\hat{m}$, $m$ are the predicted mask and ground-truth mask, respectively. $f(\cdot)$ represents any subsequent mask predictor. The backward gradients with respect to ($\theta$ or $X$) can be calculated by:
\begin{eqnarray}\label{eq8}
\frac{\partial L}{\partial (\cdot)}=\frac{\partial L}{\partial H^{\ast}}(-J_{\Phi_{\theta}}^{-1}|_{H^{\ast}})\frac{\partial F_{\theta}(H^{\ast};X)}{\partial (\cdot)},
\end{eqnarray}
where $J_{\Phi_{\theta}}^{-1}|_{H^{\ast}}$ denotes the Jacobian inverse of $\Phi_{\theta}$ at the equilibrium state $H^{\ast}$.

\subsection{Convergence Analysis}
For the convergence analysis of infinite loop presented in Eq.~\ref{eq4}, we first simulated the unrolling process by 100k steps given a random variable input and the double residual network. We calculated the norm difference between adjacent steps and found that the norm difference will gradually converge to zero. Then we calculated the spectral radius by the Jacobian matrix and the spectral radius of each step is always less than one, which guarantees the convergence of equilibrium state. Since our IFR is built on Eq.~\ref{eq1}, therefore the root of Eq.~\ref{eq1} solved will be approximately same as the equilibrium state. Given the same image, we compare the norm difference between the features generated by our IFR module and the infinite loop, where the difference between them is less than $e^{-8}$.

\subsection{Extensions on Object Detectors}
Our implicit refinement module can also be extended to object detectors with slight modifications. Notably, most one-stage object detectors tend to adopt multiple stacked convolutions in the head network to refine whole features for the final predictions~\cite{lin2017focal, tian2019fcos, duan2019centernet, zhang2020bridging}. To this end, our IFR can be directly integrated into the head network to replace the explicitly stacked convolutions (\textit{e.g.}, four consecutive $3\times 3$ convolutions involved in RetinaNet head~\cite{lin2017focal}).

Different from the implicit module plugged in Mask R-CNN, the input of implicit refinement module is each pyramidal feature $X^{'} \in R^{h_{i}\times w_{i}\times C}$ from FPN~\cite{lin2017feature}, where $h_{i}$ and $w_{i}$ are the height and width of the $i_{th}$ level feature map, respectively. The equilibrium hidden feature $H^{\ast}$ output by the implicit module is then fed into the subsequent classification and regression predictors.

\section{Experiments}
We evaluate our method on the challenging COCO~\cite{lin2014microsoft} and YouTube-VIS~\cite{yang2019video} benchmarks. We present a thorough performance comparison with baselines on both two-stage and one-stage instance segmentation approaches along with detailed ablation studies.

\subsection{Datasets and Metrics}
\noindent \textbf{COCO~\cite{lin2014microsoft}:} The COCO benchmark contains 118k images for training, 5k images for validation and 20k images for testing, involving 80 object categories with instance-level segmentation annotations. Following~\cite{he2017mask, tian2020conditional, lin2017focal, tian2019fcos}, our models are trained on \textit{train2017} set. We report the results on \textit{val2017} set for ablation study and comparison with the baselines. For evaluation, we follow the standard COCO metrics including AP, AP$_{50}$, AP$_{75}$, and AP$_{s}$, AP$_{m}$, AP$_{l}$.

\noindent \textbf{YouTube-VIS~\cite{yang2019video}:} YouTube-VIS is a large-scale video instance segmentation benchmark. It consists of 2,883 high-resolution YouTube videos, a 40-category label set, 4,883 video instances, and 131k high-quality instance masks. Following~\cite{yang2019video, cao2020sipmask}, evaluation metrics are AP, AP$_{50}$, AP$_{75}$, AR$_{1}$, AR$_{10}$.

\subsection{Implementation Details}
We employ the implicit refinement module to replace stacked convolutions in different instance segmentation frameworks. The backbone network is pretrained on ImageNet dataset while the parameters of FPN is randomly initialized. For the parameters of transformation block, we initialize them as in~\cite{he2017mask}. The hidden features $H^{0}$ are initialized to zeros. Following~\cite{he2017mask, wang2020solo, lin2017focal, tian2019fcos}, the input image from COCO is resized such that the shorter side is in the range of [640, 800] and the longer side is less or equal to 1333. In testing phase, the shorter side is set to 800. For the YouTube-VIS~\cite{yang2019video} benchmark, input size is set as $640\times360$. All the models are trained over 8 GPUs using stochastic gradient descent (SGD) with a mini-batch of 16 for 90k (1x) iterations. The initial learning rate is 0.01 and reduced by a factor of 0.1 at 60k and 80k iterations, respectively. The warm-up strategy is adopted for the first 1k iterations during the training process. The forward and backward iterations of Broyden root-solver are all set to 15. In addition, weight normalization~\cite{salimans2016weight} is adopted to stabilize the training process.

\subsection{Results on Image Instance Segmentation}
We first evaluate our IFR on the state-of-the-art two-stage and one-stage instance segmentation frameworks. 

\noindent \textbf{Two-Stage Methods:} Tab.~\ref{tab:exp_two_stage} shows the performance comparison between stacked convolutions and our IFR on two-stage instance segmentation frameworks. For all the given approaches, our IFR outperforms the stacked convolutions with less parameters. 
These results show the effectiveness of our IFR even when deploying on the state-of-the-art instance segmentation approaches, \textit{e.g.} BMask R-CNN~\cite{cheng2020boundary}. Compared to the Mask Scoring R-CNN with stacked convolutions, Mask Scoring R-CNN integrated with IFR improves +1.2\%  and +1.0\% on large (AP$_{l}$) and medium (AP$_{m}$) objects. The results reveal that our IFR can produce high-level equilibrium instance features of larger receptive field, owing to the fixed-point iteration.

\begin{figure*}[hbt]
\centering
\includegraphics[scale=0.52]{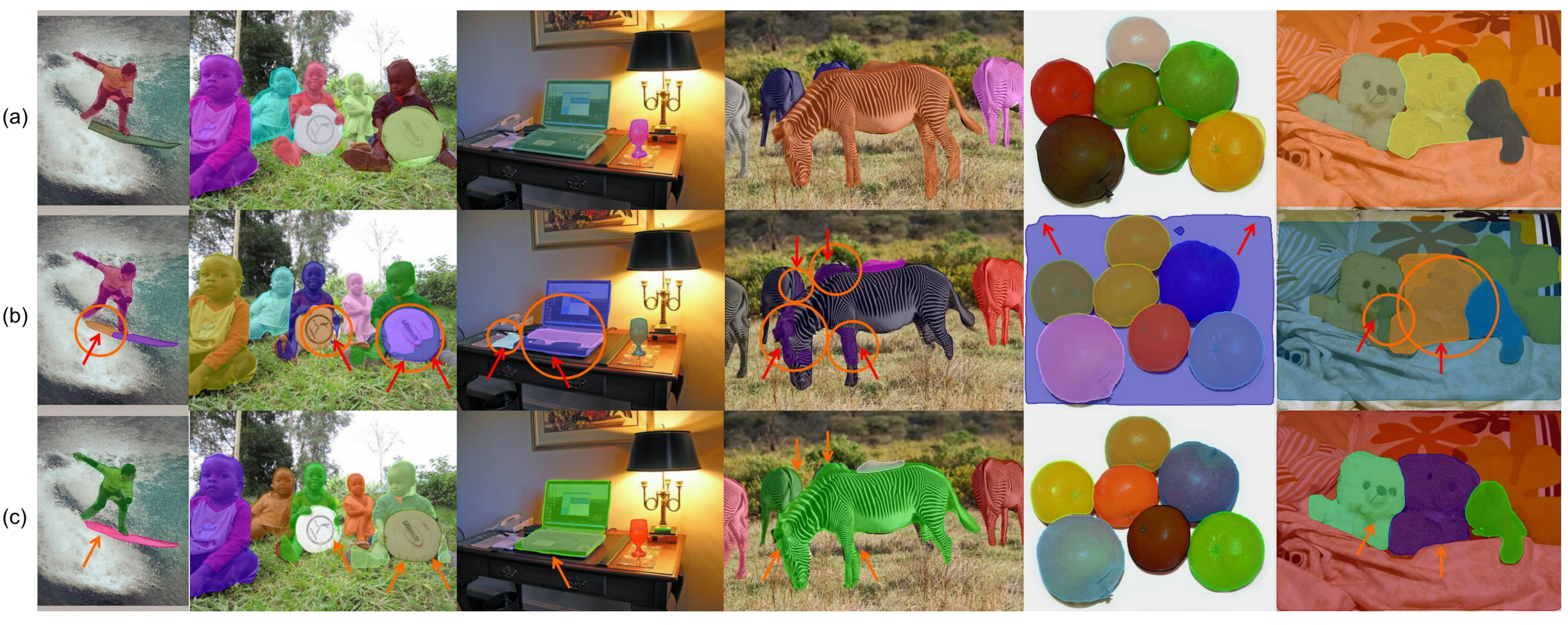}
\caption{The qualitative comparison between \textbf{(a)} ground-truth annotations, \textbf{(b)} Mask R-CNN~\cite{he2017mask}, and \textbf{(c)} improved Mask R-CNN with our implicit feature refinement on COCO \textit{val2017}.}     \label{fig3}
\end{figure*}

Fig.~\ref{fig3} illustrates the qualitative comparison on Mask R-CNN~\cite{he2017mask} between the stacked four convolutions and our proposed IFR. More visualization results are shown in Appendix A.1. Fig.~\ref{fig4} further shows the comparison between the RoI features refined by stacked convolutions and our IFR on several images from the COCO \textit{val2017} set. Specifically, Fig.~\ref{fig4} depicts the object of interest cropped from input images, original RoI features produced by RoIAlign, refined features by stacked four convolutions and our proposed IFR. From the qualitative visualization, we can easily find that the RoI features refined by the stacked convolutions are coarse and of limited receptive field. On the contrary, the RoI features refined by our IFR are much finer and of larger receptive field. It means our IFR can further enlarge the receptive field and improve semantic levels of RoI features. 

\begin{table}[h]
\caption{Performance comparison between four $3\times3$ convs and our IFR on two-stage instance segmentation frameworks, evaluated on COCO \textit{val2017} set. 1x (12 epochs) training strategy is adopted.} \vspace{-0.3cm}
\begin{center}
\resizebox{\linewidth}{!}{
\begin{tabular}{l|ccc|ccc}
\hline
Methods     &AP &AP$_{50}$ &AP$_{75}$ &AP$_{s}$ &AP$_{m}$ &AP$_{l}$ \\
\hline
Mask R-CNN  &35.1 &56.2 &37.6 &16.9 &37.4 &50.8\\
w/ IFR  &\textbf{36.1} &\textbf{56.7} &\textbf{39.0} &\textbf{17.9} &\textbf{38.5} &\textbf{51.8} \\
\hline
Cascade Mask RCNN & 36.1 & \textbf{56.7} & 38.8 & 17.0 & 38.4 & 53.1 \\
w/ IFR & \textbf{36.6} & \textbf{56.7} & \textbf{39.5} & \textbf{17.1} & \textbf{39.0} & \textbf{53.9} \\
\hline
Mask Scoring RCNN & 36.4 & 56.4 & 39.1 & 17.2 & 38.7 & 52.0 \\
w/ IFR & \textbf{36.9} & \textbf{56.6} & \textbf{40.1} & \textbf{17.4} & \textbf{39.7} & \textbf{53.2} \\
\hline
BMask R-CNN & 36.6 & 56.7 & 39.4 & \textbf{17.3} & 38.8 & 53.8 \\
w/ IFR & \textbf{37.3} & \textbf{57.0} & \textbf{40.1} & \textbf{17.3} & \textbf{39.3} & \textbf{54.5} \\
\hline
Hybrid Task Cascade & 37.3 & \textbf{58.3} & 40.5 & 19.7 & 40.4 & 51.2 \\
w/ IFR & \textbf{37.6} & 58.2 & \textbf{40.8} & \textbf{19.8} & \textbf{40.6} & \textbf{51.5} \\
\hline
\end{tabular}
}
\end{center}\vspace{-0.2cm}
\label{tab:exp_two_stage}
 \end{table}
 
\begin{figure}[hbt]
\centering
\includegraphics[scale=0.52]{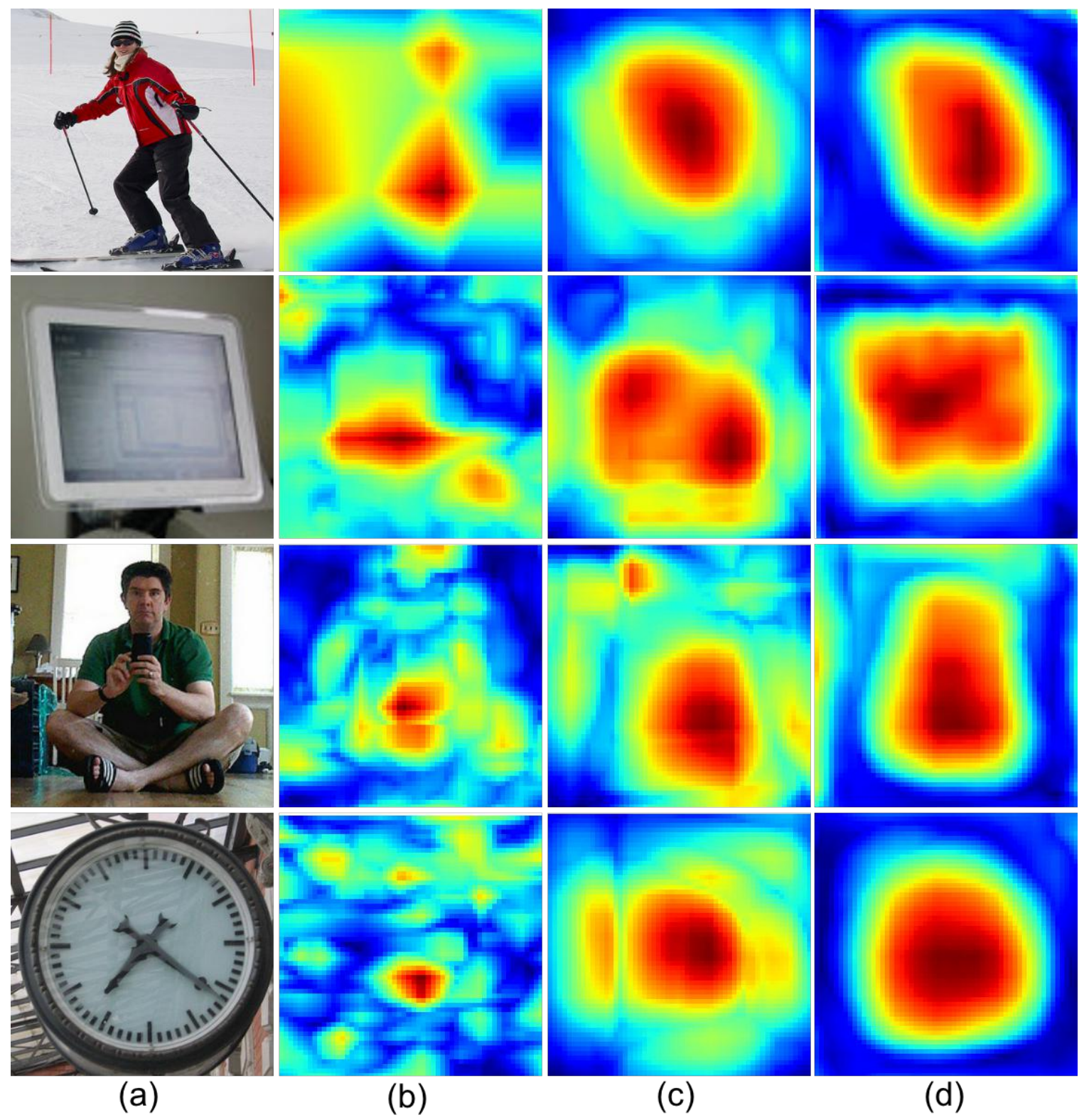}
\caption{Visualization comparison of RoI features refined by different approaches on the basis of Mask R-CNN~\cite{he2017mask}. (a) object of interest cropped from input image. (b) coarse RoI features cropped by RoIAlign before feature refinement. (c) and (d) depicts the RoI features refined by stacked convolutions and our IFR, respectively.}
\label{fig4}
\end{figure}

\noindent \textbf{One-Stage Methods:} Analogously, Tab.~\ref{Tab7} shows the performance comparison on some one-stage instance segmentation frameworks. The results demonstrate that our IFR is robust and also flexible for one-stage approaches that directly outputs the full instance masks from whole features without RoIAlign. 
For instance, our proposed IFR improves the average AP by +1.0\% mAP on MEInst~\cite{zhang2020mask}. Similarly, more improvements on large-scale objects are achieved due to the enlarged receptive field.

\begin{table}[h]
\caption{Performance comparison between four $3\times3$ convs and our IFR on one-stage instance segmentation frameworks, evaluated on COCO \textit{val2017} set. Except for SOLO~\cite{wang2020solo} and SOLOv2~\cite{wang2020solov2}, which adopt a 3x training schedule, other models are trained with 1x schedule.}\label{Tab7}
\vspace{-0.3cm}
\begin{center}
\resizebox{\linewidth}{!}{
\begin{tabular}{l|ccc|ccc}
\hline
Methods     &AP &AP$_{50}$ &AP$_{75}$ &AP$_{s}$ &AP$_{m}$ &AP$_{l}$ \\
\hline
MEInst & 31.6 & 53.5 & 33.0 & 15.0 & 34.1 & 45.3 \\
w/ IFR & \textbf{32.6} & \textbf{54.3} & \textbf{33.9} & \textbf{15.9} & \textbf{35.1} & \textbf{47.3} \\
\hline
BlendMask & 35.8 & 56.3 & 38.2 & 17.3 & 39.1 & 50.8 \\
w/ IFR & \textbf{36.4} & \textbf{56.7} & \textbf{39.0} & \textbf{17.6} & \textbf{39.3} & \textbf{52.3} \\
\hline
SOLO & 35.3 & 56.6 & 37.2 & \textbf{16.0} & 38.1 & 52.2 \\
w/ IFR & \textbf{35.9} & \textbf{56.7} & \textbf{38.2} & 14.6 & \textbf{39.2} & \textbf{53.3} \\
\hline
SOLOv2 & 37.3 & 57.5 & 39.7 & 15.2 & 40.8 & 56.5 \\
w/ IFR & \textbf{37.6} & \textbf{57.8} & \textbf{39.9} & \textbf{16.0} & \textbf{41.2} & \textbf{57.3} \\
\hline
CondInst & 35.5 & 56.2 & 37.8 & 16.7 & 39.1 & 50.8 \\
w/ IFR & \textbf{36.0} & \textbf{56.6} & \textbf{38.3} & \textbf{17.2} & \textbf{39.2} & \textbf{52.0} \\
\hline
BoxInst & 30.5 & 52.3 & 30.8 & 13.5 & \textbf{33.4} & 45.1 \\
w/ IFR & \textbf{30.9} & \textbf{52.7} & \textbf{31.5} & \textbf{14.3} & 33.2 & \textbf{46.5} \\
\hline
\end{tabular}
}
\end{center}\vspace{-0.2cm}
\label{tab:exp_one_stage}
 \end{table}

\subsection{Results on Video Instance Segmentation}
We also evaluate the effectiveness of our implicit feature refinement on video instance segmentation. Tab.~\ref{Tab8} shows the performance comparison between stacked $3\times3$ convolutions and our IFR on two state-of-the-art video instance segmentation frameworks. MaskTrack R-CNN~\cite{yang2019video} is a typical two-stage method while SipMask~\cite{cao2020sipmask} is an one-stage framework. The experimental results demonstrate that IFR is also effective for video instance segmentation since a strong and robust feature representation is generated by the fixed-point iteration. Besides, our IFR provides relatively larger gains under multi-scale training setting. It improves both the MaskTrack R-CNN~\cite{yang2019video} and SipMask~\cite{cao2020sipmask} by +0.9\% in terms of mask accuracy.

 \begin{figure*}[]
\centering
\includegraphics[scale=0.50]{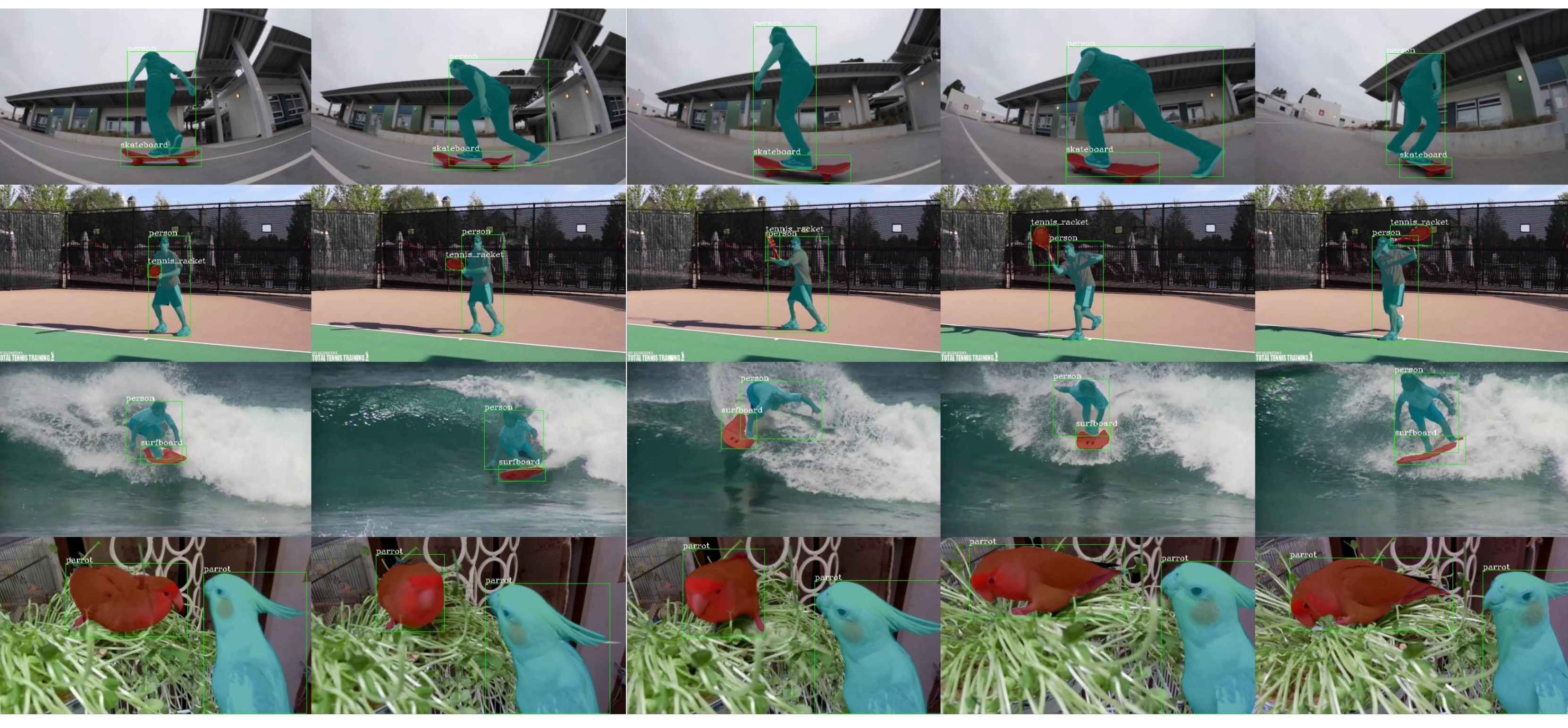}
\caption{Qualitative results of IFR enhanced SipMask~\cite{cao2020sipmask} on several example frames from different videos in YouTube-VIS~\cite{yang2019video} validation set. The object with same predicted identity has the same color.} \label{fig5}
\end{figure*}

\begin{table}[hbt]
\caption{Performance comparison between four $3\times3$ convs and our IFR on video instance segmentation frameworks, evaluated on YouTube-VIS validation set. $"$MST$"$ refers to multi-scale training. All the models are reproduced by default setting reported in~\cite{yang2019video, cao2020sipmask} and trained with a 1x training schedule.}\label{Tab8}
\vspace{-0.3cm}
\begin{center}
\resizebox{\linewidth}{!}{
\begin{tabular}{l|c|ccc|cc}
\hline
Methods & MST &AP &AP$_{50}$ &AP$_{75}$ &AR$_{1}$ &AR$_{10}$ \\
\hline
MaskTrack R-CNN & \multirow{2}{*}{-} & 29.9 & 50.4 & 31.8 & \textbf{31.6} & \textbf{35.9} \\
w/ IFR & & \textbf{30.5} & \textbf{51.3} & \textbf{32.2} & 31.0 & 35.7 \\
\hline
MaskTrack R-CNN & \multirow{2}{*}{\checkmark} & 30.5 & 50.7 & \textbf{33.2} & 31.6 & 35.9 \\
w/ IFR & &\textbf{31.4} & \textbf{53.3} & 32.9 & \textbf{32.0} & \textbf{37.4} \\
\hline
SipMask & \multirow{2}{*}{-} & 31.6 & 51.2 & 33.5 & 33.2 & 37.7 \\
w/ IFR & & \textbf{31.9} & \textbf{53.0} & \textbf{33.6} & \textbf{33.9} & \textbf{39.6} \\
\hline
SipMask & \multirow{2}{*}{\checkmark} & 33.7 & 52.6 & \textbf{36.9} & \textbf{35.1} & \textbf{40.7} \\
w/ IFR & & \textbf{34.6} & \textbf{55.6} & 36.7 &
\textbf{35.1} & 40.4 \\
\hline
\end{tabular}
}
\end{center}\vspace{-0.2cm}
\label{tab:exp_vis}
 \end{table}

Fig.~\ref{fig5} shows the qualitative results of IFR enhanced SipMask on several frames from the YouTube-VIS validation set~\cite{yang2019video}. As illustrated, our method can accurately segment instances in each frame and associates them across frames. More visualization results are shown in Appendix A.2.

\subsection{Results on Object Detection}
We further extend IFR to the detection head of some one-stage object detectors. Tab.~\ref{Tab9} shows the performance comparison between the stacked convolutions and our IFR on one-stage object detectors. Compared with stacked convolutions, our IFR achieves better performance with less parameters, especially on large-scale object detection (AP$_{l}$). The results demonstrate that our IFR can produce equilibrium features of global receptive field. Therefore, our IFR can serve as a strong alternative of explicitly stacked convolutions in many object recognition tasks.

\begin{table}[hbt]
\caption{Performance comparison between four $3\times3$ convs and our IFR on object detectors, evaluated on COCO \textit{val2017} set. 1x training strategy is adopted. For the given object detectors, our IFR outperforms the four $3\times3$ convolutions with less parameters.}\label{Tab9}
\vspace{-0.3cm}
\begin{center}
\resizebox{\linewidth}{!}{
\begin{tabular}{l|ccc|ccc}
\hline
Methods     &AP &AP$_{50}$ &AP$_{75}$ &AP$_{s}$ &AP$_{m}$ &AP$_{l}$ \\
\hline
RetinaNet &36.3 &56.1 &39.1 &21.3 &40.1 &47.9 \\
w/ IFR  &\textbf{36.8} &\textbf{57.4} &\textbf{39.6} &\textbf{21.4} &\textbf{40.5} &\textbf{48.1} \\
\hline
FreeAnchor & 38.4 & 57.0 & 41.1 & 21.9 & 41.7 & 51.8 \\
w/ IFR & \textbf{39.5} & \textbf{58.5} & \textbf{42.3} & \textbf{22.4} & \textbf{42.3} & \textbf{53.7} \\
\hline
FCOS  &38.6 &57.8 &41.7 &23.2 &42.4 &49.7 \\
w/ IFR    &\textbf{39.0} &\textbf{57.9} &\textbf{42.1} &\textbf{23.4} &\textbf{42.5} &\textbf{50.7} \\
\hline
RepPoints & 38.5 & 58.8 & 41.5 & 22.4 & \textbf{42.5} & 51.1 \\
w/ IFR & \textbf{38.8} & \textbf{58.9} & \textbf{41.8} & 21.9 & \textbf{42.5} & \textbf{52.1} \\
\hline
\end{tabular}
}
\end{center}\vspace{-0.2cm}
\label{tab:dethead}
 \end{table}

\vspace{-0.1cm}
\subsection{Ablation Study}
\noindent \textbf{Different Refinement Strategies:} We first analyse the impact of different feature refinement strategies (explicit vs implicit). As presented in Eq.~\ref{eq2}, the explicit feature refinement strategy stacks four weight-independent blocks to refine RoI features. For the implicit refinement strategy, both the unrolling and Broyden solvers can be employed. The unrolling solver stacks the blocks in a weight-sharing manner while the Broyden solver directly produces the equilibrium feature via the fixed point iteration. For fair comparison, double residual network is adopted as the basic block for different refinement strategies. Tab.~\ref{Tab2} shows the performance comparison on Mask R-CNN~\cite{he2017mask}, where the unrolling process iterates four blocks with or without weight sharing. The implicit unrolling strategy achieves the same performance (36.0 AP) compared to the explicit strategy with only 30.0\% parameters in the mask head. In contrast, implicit feature refinement with the Broyden solver produces 36.1 AP without explicitly iterating these unrolling blocks, since it can directly solve out the fixed point of implicit model constructed.

\begin{table}[hbt]
\caption{Analyzing the impact of different refinement strategies with the unrolling and Broyden solvers. "P(M)" indicates the number of parameters in the mask head network of Mask R-CNN~\cite{he2017mask}.}\label{Tab2}
\begin{center}
\resizebox{\linewidth}{!}{
\begin{tabular}{l|c|ccc|ccc|c} 
\toprule
Strategy & Solver &AP &AP$_{50}$ &AP$_{75}$ &AP$_{s}$ &AP$_{m}$ &AP$_{l}$ & $P(M)$\\
\midrule
\multirow{1}{*}{Explicit} & Unrolling & 36.0 & 56.7 & 38.9 & 17.4 & 38.5 & 51.7 & 5.0 \\
\midrule
\multirow{2}{*}{Implicit} & Unrolling & 36.0 & 56.6 & 38.9 & 17.4 & 38.3 & 51.6 & \textbf{1.5} \\
& Broyden & \textbf{36.1} & \textbf{56.7} & \textbf{39.0} & \textbf{17.9} & \textbf{38.5} & \textbf{51.8} & \textbf{1.5} \\
\bottomrule
\end{tabular}
}
\end{center}
\label{tab:stacking_block_number}
\end{table}

\noindent \textbf{Iterations of the Broyden Solver:} As mentioned above, we employ the Broyden solver~\cite{bai2019deep} to obtain the fixed point. For optimization, the iteration number of the solver is a hyper-parameter. Tab.~\ref{Tab3} shows the continuous improvements as the number of iterations increases from 3 to 15 while no further improvement if increased to 20. Considering both efficiency and accuracy, the iterations of Broyden solver are set to 15 for both forward and backward propagation.

\begin{table}[H]
\caption{Impact of the iterations of Broyden solver.}\label{Tab3}
  \centering
  \resizebox{0.7\linewidth}{!}{
  \begin{tabular}{l|ccccc}
    \hline
Num. of iters & 3 & 5 & 10 & 15 & 20\\
    \hline
    $AP$    &31.7 &33.2 &35.9  & \textbf{36.1} & \textbf{36.1} \\
    \hline
  \end{tabular}\vspace{-0.1cm}
  }
  \label{tbl:iters_broyden}
\end{table}

\noindent \textbf{Double Residual Connection:} We introduce the double residual connections for the explicit design of nonlinear transformation. Here, we evaluate its effect on Mask R-CNN~\cite{he2017mask}. Specifically, we remove these two residual connections in our IFR and only keep two consecutive convolution layers. Tab.~\ref{Tab4} show that double connection improves the overall performance since it benefits from residual learning and smooth gradient propagation.

\begin {table}[hbt]
\centering
\caption{Impact of integrating the double residual connections into the standard Mask R-CNN on COCO \textit{val2017} set.}\label{Tab4}
\vspace{-0.2cm}
\resizebox{\linewidth}{!}{%
\begin{tabular}{c|ccc|cccc}
\hline
Double res-connection  &AP &AP$_{50}$ &AP$_{75}$ &AP$_{s}$ &AP$_{m}$ &AP$_{l}$\\
\hline
  -  & 35.8 & 56.6 & 38.6 & 17.2 & 38.2 & 51.1 \\
  \checkmark & \textbf{36.1} & \textbf{56.7} & \textbf{39.0} & \textbf{17.9} & \textbf{38.5} & \textbf{51.8} \\
\hline
\end{tabular}
}\vspace{0.2cm}
\label{ablation_combine}
\end{table}

\noindent \textbf{Number of Intermediate Channels in Res-Block:} For the employed res-block, we evaluate the effect of intermediate channels between two $3\times$3 convolutions. Tab.~\ref{Tab5} shows the performance comparison on Mask R-CNN~\cite{he2017mask}, where $X_{1}$ denotes the basic number of channels, 256. The results show continuous improvements as the number of intermediate channels increases from 32 to 256. To achieve the trade-off between performance and parameter burden, we set the number of intermediate channels to 256.

\begin{table}[h]
\caption{Impact of the number of convolution channels.}\label{Tab5}
  \centering
  \resizebox{0.9\linewidth}{!}{
  \begin{tabular}{l|ccccc}
    \hline
Num. of channels & $X_{1/8}$ & $X_{1/4}$ & $X_{1/2}$ & $X_{1}$ & $X_{2}$ \\
    \hline
    $AP$ & 35.3 & 35.4 & 35.6 & \textbf{36.1} & \textbf{36.1} \\
    $P(M)$ & 0.4 & 0.6 & 0.9 & 1.5 & 2.6 \\
    \hline
  \end{tabular}\vspace{-0.1cm}
  }
  \label{tbl:num_channels}
\end{table}

\noindent \textbf{Larger Backbone:} To further demonstrate the effectiveness of our IFR, we also conduct the experiments with larger backbones. Tab.~\ref{Tab10} shows that our IFR can still produce better performance than stacked convolutions even when using larger backbone.

\begin{table}[h]
\caption{Performance comparison between the stacked four $3\times3$ convolutions and our IFR on Mask R-CNN~\cite{he2017mask} with larger backbone. 3x (36 epochs) training strategy is adopted.}\label{Tab10}
\vspace{-0.3cm}
\begin{center}
\resizebox{\linewidth}{!}{
\begin{tabular}{l|ccc|ccc}
\hline
Methods     &AP &AP$_{50}$ &AP$_{75}$ &AP$_{s}$ &AP$_{m}$ &AP$_{l}$ \\
\hline
Mask R-CNN-Res101 &38.6 &\textbf{60.5} &41.4 &19.1 &41.2 &55.3 \\
w/ IFR    &\textbf{39.0} &\textbf{60.5} &\textbf{42.0} &\textbf{19.9} &\textbf{41.8} &\textbf{56.0} \\
\hline
\end{tabular}
}
\end{center}\vspace{-0.2cm}
 \end{table}

\section{Conclusion}
In this paper, we propose an implicit feature refinement framework for image/video instance segmentation. Current instance segmentation methods tend to apply multiple convolutions to refine instance features but the refined features are of limited receptive field. In this paper, we propose to refine instance features of global receptive field via a simulated infinite-depth network, which can be employed in both one-stage and two-stage approaches. The proposed IFR produces improved performance on most state-of-the-art frameworks while reducing the overall parameter burden.

\bibliographystyle{ACM-Reference-Format}
\balance
\bibliography{sample-base}

\end{document}